\PassOptionsToPackage{table}{xcolor}
\PassOptionsToPackage{comma,numbers,sort,compress}{natbib}
\documentclass[11pt,letterpaper,logo]{googledeepmind}
\usepackage[comma,numbers,sort,compress]{natbib}
\usepackage{dm-colors}
\usepackage{verbatim}
\usepackage{etoolbox}
\usepackage{xspace}
\usepackage{subcaption}
\usepackage{multirow}
\usepackage[capitalize]{cleveref}
\crefname{section}{Sec.}{Secs.}
\Crefname{section}{Section}{Sections}
\crefname{table}{Tab.}{Tabs.}
\Crefname{table}{Table}{Tables}

\titleformat{\paragraph}[runin]{\normalfont\normalsize\bfseries}{}{0pt}{#1}
\titlespacing*{\paragraph}{0pt}{5pt plus 1pt minus 1pt}{1em}
\pretocmd{\section}{\Needspace{6\baselineskip}}{}{}
\pretocmd{\subsection}{\Needspace{6\baselineskip}}{}{}

\fancypagestyle{firststyle}{%
  \fancyhf{}%
  \fancyhead[L]{\includegraphics[width=90pt]{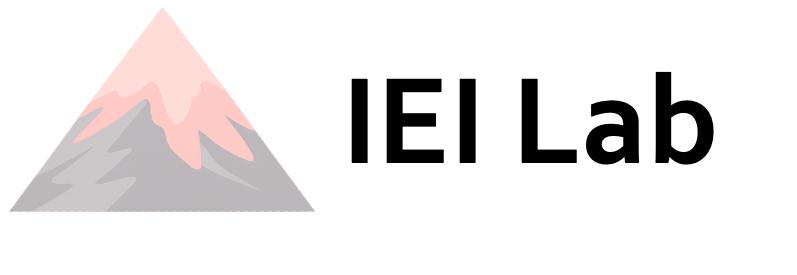}\vspace{-12pt}}%
  \fancyfoot[R]{\footerfont\thepage}%
}
\newsavebox{\successratebox}

\usepackage{wrapfig}

\title{Dr. Grasp-RL: Diverse and Robust Dexterous Grasping via Reinforcement Learning}
\author[1]{Kunqi Xu}
\author[1]{Xiaokang Sun}
\author[1]{Xueyan Zou}
\affil[1]{Tsinghua University - IEI Lab}

\affil[ ]{\protect\\[-0.2em]
\href{https://xkq1qkx.github.io/Diverse-Manip-RL-web/}{https://xkq1qkx.github.io/Diverse-Manip-RL-web/}}

\hypersetup{
  pdftitle={Dr. Grasp-RL: Diverse and Robust Dexterous Grasping via Reinforcement Learning},
  pdfauthor={Kunqi Xu, Xiaokang Sun, Xueyan Zou},
  colorlinks=true,
  citecolor=magenta
}
\begin{abstract}
Human interaction with physical objects relies on a combination of dexterity and robustness. In robotic manipulation, dexterity requires not only flexible end effectors but also a diverse repertoire of manipulation strategies, while robustness requires closed-loop adaptation to disturbances and unexpected situations. Two prominent research directions in dexterous grasping have largely addressed these capabilities separately. Generative approaches model multimodal distributions to produce diverse grasp poses, but their reliance on open-loop execution can limit robustness. Reinforcement learning approaches learn closed-loop grasping policies, but can converge to a narrow range of successful grasping strategies, limiting behavioral diversity. We introduce a time-reversed curriculum with a coordinated reset mechanism that enables a simple MLP policy trained with
PPO to capture multiple modes of the grasp-pose distribution while maintaining closed-loop control. Experiments demonstrate both diverse grasping behaviors and
robust closed-loop execution. Trained on only approximately 400 objects, our policy further demonstrates strong generalization when evaluated on more than 20,000 unseen objects.
\end{abstract}

\begin{document}
\maketitle

\begin{figure}[htbp]
  \centering
  \includegraphics[width=\linewidth]{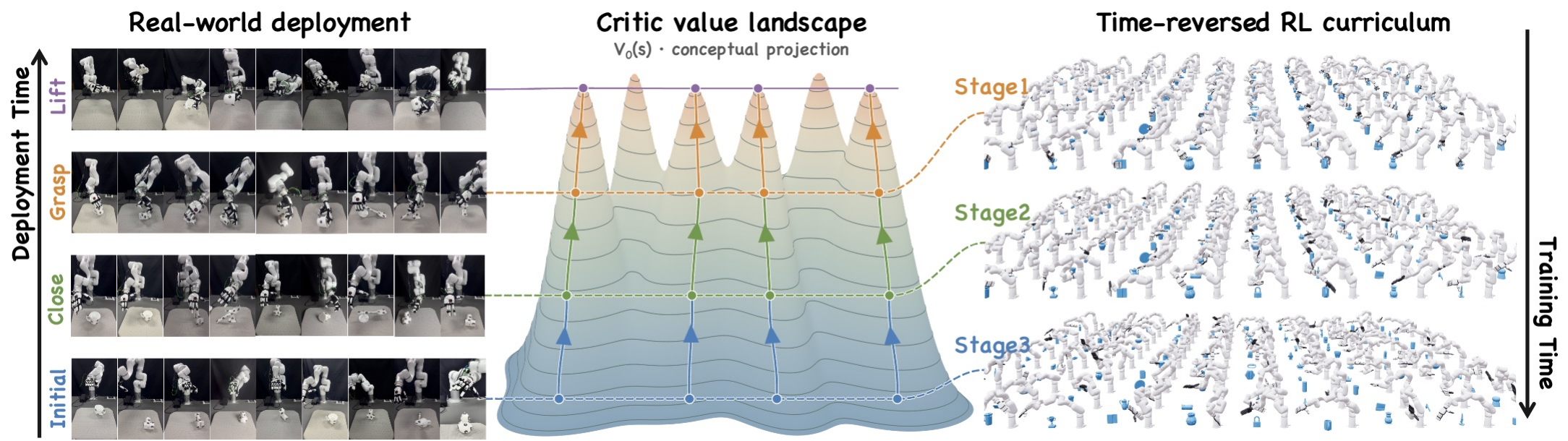}
  \caption{\textbf{Diverse and robust grasping through RL.}
Conceptually, multiple peaks in the critic value landscape support diverse grasping modes, while dense, smooth guidance supports closed-loop control and recovery. PPO jointly learns the critic and policy.}
  \label{fig:teaser}
\end{figure}

\section{Introduction}
\label{sec:intro}

What is the essence of human's ability to manipulate objects? How can we make robots exhibit human-level intelligence in their object manipulation performance? This remains an unsolved and even poorly understood problem~\cite{doi:10.1126/science.aat8414,kroemer2020reviewrobotlearningmanipulation}. Humans can grasp the same object in many different ways, adapting their wrist pose and finger configuration to the current situation~\cite{7243327}. This flexibility suggests a desirable capability for robotic grasping: a single policy should support multiple effective grasping modes and select suitable actions from the current hand–object state. Diversity and feedback are complementary~\cite{kumar2020solutionneedfewshotextrapolation,chen2026dexxlearningvisualtactiledexterous}. Alternative grasping modes provide options when an approach is unfavorable, while closed-loop control allows the robot to adjust its actions and retry after failed contact~\cite{kalashnikov2018qtoptscalabledeepreinforcement}. The challenge is to learn a policy that retains this repertoire while realizing its behaviors through continuous physical interaction.

A useful perspective on this challenge comes from the distinction
between evaluating candidate solutions and generating them~\cite{florence2021implicitbehavioralcloning,Bohg_2014}.
Generative modeling offers an expressive means of representing
multiple plausible solutions, as illustrated by advances in image
and video generation~\cite{ho2020denoisingdiffusionprobabilisticmodels,ho2022videodiffusionmodels}.
In dexterous grasping, grasp-synthesis approaches have similarly explored
large-scale grasp data and diverse grasp-pose distributions.
However, a grasp proposal alone does not specify how to reach, maintain, or recover it through physical interaction.
Conversely, some policy-learning approaches integrate grasping more
directly with feedback close-loop control, either through reinforcement
learning or subsequent policy distillation.
Yet optimizing task success does not necessarily encourage a
policy to preserve multiple successful grasping modes:
learning can concentrate on a narrow repertoire of effective
behaviors~\cite{eysenbach2018diversityneedlearningskills,haarnoja2017reinforcementlearningdeepenergybased}.
Despite progress along both directions, jointly learning diverse
grasping strategies and reliable close-loop execution remains
a central challenge.

We revisit this challenge from a value-function perspective.
Rather than explicitly generating a target grasp pose, we seek to
learn a state-value landscape---a
scalar field over grasping states---with two key properties:
multiple high-value peaks and dense, informative guidance.
Each peak should correspond to a distinct effective grasp
configuration, allowing the landscape to represent multiple
valid ways of grasping the same object.
Far from the object, the value should generally increase as the
hand approaches a feasible grasp from different directions;
near contact, it should distinguish effective wrist poses and
finger configurations according to the object's geometry.
Importantly, this guidance should extend beyond successful
grasp configurations to the intermediate states encountered
during approach, contact, and recovery.
Intuitively, a controller could then select actions that produce
the greatest expected increase in value under the robot's
dynamics, repeatedly updating its decisions from the current
state.
Multiple peaks would preserve alternative grasping strategies,
while dense value guidance would support closed-loop execution
and adaptation after failures or disturbances.

This intuition naturally motivates an actor--critic approach~\cite{NIPS1999_6449f44a}:
the critic learns to evaluate states associated with different
grasping modes, while the actor learns feedback actions that
lead toward high-return outcomes.
We therefore introduce Dr.~Grasp-RL, which uses PPO~\cite{schulman2017proximalpolicyoptimizationalgorithms} with simple
multilayer perceptron (MLP) networks to learn diverse, closed-loop
dexterous grasping.
Through a time-reversed training curriculum and diverse,
instructive reset states, we progressively extend learning from
states near successful grasps to earlier and more challenging
stages of the interaction.
We just use a simple reward proportional to object lift height.
Our experiments demonstrate diverse grasping behaviors, robust
closed-loop execution, and strong generalization across object
geometries. Our contributions are threefold:
\begin{itemize}
    \item We introduce Dr.~Grasp-RL, a framework for learning a
    diverse repertoire of dexterous grasping behaviors within a
    single closed-loop policy using MLP-based PPO.

    \item We develop a time-reversed curriculum coupled with
    diverse, curriculum-aligned reset states to facilitate
    learning across multiple grasping modes with a simple
    lift-height reward, without requiring an explicit
    grasp-pose generator at execution time.

    \item We evaluate the learned policy on more than 20,000
    unseen objects after training on no more than 420 objects,
    and analyze its grasp diversity, closed-loop recovery
    behaviors, and generalization.
\end{itemize}
\section{Related Work}
\label{sec:related_work}

\paragraph{Grasp generation and demonstration editing.}
Generative grasp models provide diverse proposals as priors for
downstream planning and execution.
Optimization-based synthesis supplies grasp datasets such as
DexGraspNet~\cite{wang2023dexgraspnetlargescaleroboticdexterous} and the taxonomy-conditioned
Dexonomy~\cite{chen2025dexonomysynthesizingdexterousgrasp}.
GenDexGrasp~\cite{li2023gendexgraspgeneralizabledexterousgrasping} uses hand-agnostic contact maps,
while DexGraspNet 2.0~\cite{zhang2024dexgraspnet20learninggenerative} and
DexGrasp Anything~\cite{zhong2025dexgraspanythinguniversalrobotic}
extend diffusion-based synthesis to cluttered scenes and
physics-aware grasp generation, respectively.
Dex1B~\cite{ye2025dex1blearning1bdemonstrations} scales generative demonstrations for
grasping and articulation.
At the trajectory level,
DemoGrasp~\cite{yuan2025demograspuniversaldexterousgrasping} and
DemoFunGrasp~\cite{mao2025universaldexterousfunctionalgrasping}
use RL-based demonstration editing, with the latter conditioning
on grasp styles and affordances.
Both also distill edited demonstrations into visual policies.
However, generating grasp or trajectory proposals separately
from execution decouples planning from ongoing physical interaction.
Without feedback-driven proposal updates or replanning, fixed
targets can become unsuitable under object motion or unexpected
contact, limiting robustness and real-time adaptation.
Our approach learns grasping within a single feedback policy, allowing actions to adapt continuously to the
evolving hand-object state.

\paragraph{Closed-loop dexterous grasping and policy learning.}
DexMV~\cite{qin2022dexmvimitationlearningdexterous} and DexVIP~\cite{mandikal2022dexviplearningdexterousgrasping} use human demonstrations and hand-pose priors to guide policy learning. UniDexGrasp~\cite{xu2023unidexgraspuniversalroboticdexterous} couples grasp proposals with a goal-conditioned RL controller, while UniDexGrasp++~\cite{wan2023unidexgraspimprovingdexterousgrasping} combines geometry-aware curricula with iterative generalist--specialist learning. UniGraspTransformer~\cite{wang2025unigrasptransformersimplifiedpolicydistillation} distills object-specific experts into a universal Transformer policy. RobustDexGrasp~\cite{zhang2025robustdexgrasprobustdexterousgrasping} combines hand-centric geometry with teacher--student learning, and ResDex~\cite{huang2025efficient} uses residual learning and mixture-of-experts to support different grasping styles. These approaches enable feedback control, often through specialized priors, expert construction, or multi-stage learning. Nevertheless, closed-loop task success does not itself ensure broad coverage of effective grasping modes: return maximization can concentrate learning on a limited repertoire of successful behaviors. We address this exploration challenge through diverse resets and a time-reversed curriculum, while keeping the policy architecture and reward simple. Our approach learns multiple grasping modes within a single policy that adapts its actions through continuous feedback.

\begin{figure}[tbp]
    \centering
    \includegraphics[width=\textwidth,height=.72\textheight,keepaspectratio]{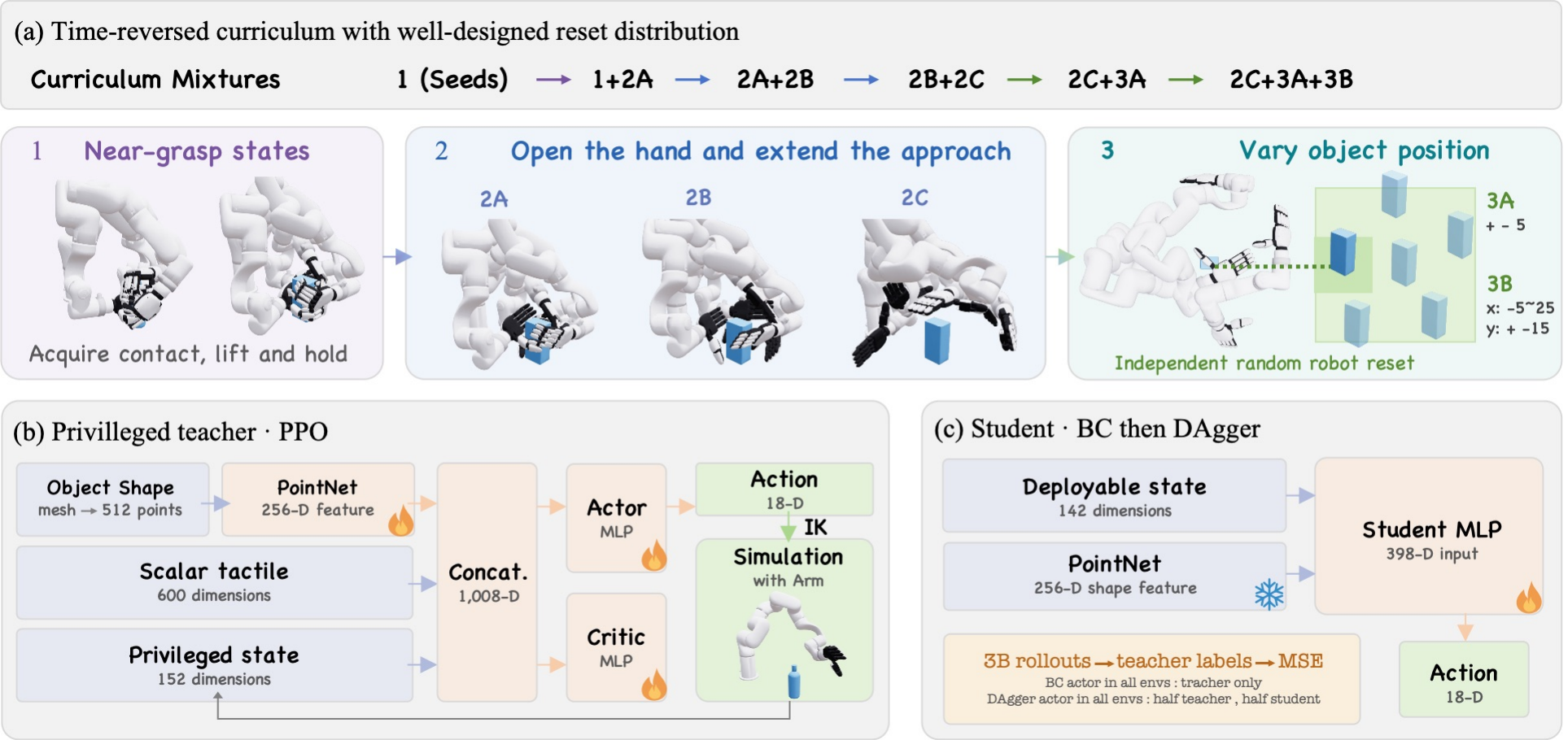}
    \caption{\textbf{Overview of Dr.~Grasp-RL.}
    \textbf{(a)} Training progresses from diverse near-grasp resets to
    open-hand approach states and independent object-position randomization.
    Earlier reset groups remain in later training mixtures.
    \textbf{(b)} PPO jointly trains a shape encoder and an MLP actor--critic
    using privileged observations and a lift-height reward. \textbf{(c)} BC followed by mixed-rollout DAgger
    trains a smaller-input student with a copied, frozen shape encoder.}
    \label{fig:method_overview}
\end{figure}

\paragraph{Curricula, reset distributions, and exploration in RL.}
Reverse Curriculum Generation~\cite{florensa2018reversecurriculumgenerationreinforcement} expands
initial states outward from goals, while
Backplay~\cite{resnick2022backplaymanmussimmer} moves initialization backward
along demonstrations.
Go-Explore~\cite{ecoffet2021goexplorenewapproachhardexploration} returns to promising archived
states before further exploration, and
Goal GAN~\cite{florensa2018automaticgoalgenerationreinforcement} generates goals of appropriate
difficulty.
More recently, OmniReset~\cite{yin2026emergentdexteritydiverseresets} uses diverse,
programmatically generated resets to scale manipulation
without an explicit curriculum.
A Balanced Data Diet~\cite{zhang2026balanceddatadietaddressing} allocates training
samples near the policy's competence frontier; related
difficulty adaptation also appears in RLVE~\cite{zeng2026rlvescalingreinforcementlearning}
for language models.
These studies primarily use curricula and reset distributions
to improve exploration and broaden task coverage.
Beyond discovering successful solutions, we investigate how
these mechanisms can help a single policy acquire and retain
multiple effective grasping modes for the same object.
This is particularly challenging in dexterous grasping, where
a single policy must coordinate wrist motion and finger contacts
across distinct grasp configurations.

\section{Method}
\label{sec:method}

Dr.~Grasp-RL learns diverse grasping behaviors by organizing the states
from which reinforcement learning begins. We first expose a policy to
multiple near-grasp configurations, then progressively move initialization
toward earlier interaction stages: opening the fingers, withdrawing the
hand, and varying the object's position independently of the robot.
The same policy learns to complete the interaction from each distribution.
Figure~\ref{fig:method_overview} summarizes this time-reversed curriculum
and the subsequent distillation into a policy with deployable inputs.

\subsection{Task and Feedback Control}
\label{sec:method_task}

We consider tabletop grasping with a seven-joint xArm7 and an XHand
with 12 actuated finger joints. The policy acts at 10\,Hz in mjlab
~\cite{zakka2026mjlablightweightframeworkgpuaccelerated}
running at 1,000\,Hz, with a maximum episode duration of 20\,s.
Let $s_t$ denote the simulator state and $o_t$ the policy observation.
The normalized command $a_t\in[-1,1]^{18}$ contains three hand-frame
translation increments, three local rotation increments, and 12 finger
joint increments. Their per-axis scales are 1\,cm, $10^\circ$, and
0.2\,rad, respectively. An inverse-kinematics controller maps the desired
hand-root pose to arm joint targets; finger targets are computed from
the measured joint angles and constrained by joint and target-rate limits.
The policy observes the resulting state before issuing the next command.
The same action interface is used throughout training and deployment.

\subsection{Time-Reversed Curriculum with Diverse Resets}
\label{sec:method_curriculum}

\paragraph{Learning backward through initialization.}
A successful grasp can be reached through different combinations of
approach direction, wrist orientation, and finger configuration.
Exploring all of these possibilities from distant initial states is
difficult when rewards depend only on lifting the object.
We therefore begin near diverse candidate grasps and gradually extend the
region from which the policy must succeed. This exposes multiple grasping
modes to reward early in training, before requiring longer approaches.
At curriculum level $\ell$, parallel environments are assigned to a
mixture of reset groups:
\begin{equation}
    \rho_{\ell}(s_0)=\sum_{g\in\mathcal G_{\ell}}
    w_{\ell,g}\rho_g(s_0),
    \qquad \sum_g w_{\ell,g}=1.
    \label{eq:reset_mixture}
\end{equation}
Here $w_{\ell,g}$ describes the fraction of environments allocated to
group $g$; completed episodes need not occur in these proportions.
We change these allocations and initial states while keeping the
observation, action, network, and reward definitions fixed.
\paragraph{Stage 1: diverse near-grasp initialization.}
We construct a bank of robot--object configurations primarily from
Dexonomy grasp templates~\cite{chen2025dexonomysynthesizingdexterousgrasp}.
Each parallel environment is tied to one fixed reset configuration.
Objects begin at their nominal positions, and the hand starts near a
candidate grasp. The policy learns to acquire contact, establish a grasp,
lift the object, and maintain its height.
Different templates expose the learner to different feasible interaction
patterns without a target-pose tracking or grasp-type reward.
The subsequent curriculum and distillation use a common set of
418 objects and 9,036 Dexonomy-derived seed configurations.
The supplementary material gives the initial-bank composition,
filtering criteria, and complete training allocations.

\paragraph{Stage 2: opening the hand and extending the approach.}
We derive three reset groups of increasing approach difficulty.
In \textbf{2A}, finger angles interpolate halfway from the original grasp
toward a predefined open-hand posture,
\begin{equation}
    q^{\mathrm{2A}}=\tfrac12q^{\mathrm{grasp}}
                    +\tfrac12q^{\mathrm{open}}.
\end{equation}
In \textbf{2B}, the hand is fully open and withdrawn by 2--8\,cm along
the original approach direction from a corrected, feasible open-hand
reference pose. In \textbf{2C}, each reset samples uniformly from a
shared library of 200 open-hand robot configurations with more varied
approach directions and orientations. This library is checked against
all 418 objects. Its hand--object separation, measured as a projected
collision-geometry gap along the object-to-palm ray, is 5--25\,cm.
Geometric screening checks clearance, self-collision, joint limits,
and IK consistency at the reset endpoint.

The three training levels mix the original seed configurations with 2A,
then 2A with 2B, and finally 2B with 2C. The first two transitions require
the newly introduced group's mean success rate to reach 60\%, with at
least ten completed episodes for every reset. We average per-reset
success rates over rolling windows of at most 20 episodes, preventing
quickly completed easy resets from dominating the promotion statistic.
The final level is trained further without another automatic threshold.

\paragraph{Stage 3: independent object-position variation.}
We next decouple object position from robot initialization.
Group \textbf{3A} samples object offsets uniformly within
$[-5,5]\times[-5,5]$\,cm, while \textbf{3B} expands this range to
$[-5,25]\times[-15,15]$\,cm. Positive $x$ points away from the arm base.
Each group has its own validated 200-pose robot library.
Robot configurations and object offsets are sampled independently:
the hand does not translate with the object.
The wider 3B region can require more distant robot starts than 2C.

Following a 3A-only warm start, the first mixed level allocates
2C and 3A environments in a $3{:}20$ ratio. Once the equal-object mean
3A success rate reaches 70\%, the second level mixes 2C, 3A, and 3B in
a $1{:}5{:}17$ ratio. This statistic uses up to 100 recent episodes per
object and requires at least ten episodes for every object.
Retaining earlier groups continues to exercise previously learned
behaviors as positional variation increases.
No auxiliary behavior-preservation loss is added.

\subsection{Privileged Actor--Critic Learning}
\label{sec:method_teacher}

\paragraph{Observations and architecture.}
The teacher receives a 152-dimensional state vector containing relative
object and hand poses, robot proprioception, object dynamics and geometry,
contact summaries, and episode progress, together with 600 scalar tactile
readings. A PointNet-style encoder~\cite{qi2017pointnetdeeplearningpoint} maps 512 points
sampled from the object's complete mesh surface to a 256-dimensional
shape feature. Points are expressed in the object frame, centered by
the mesh bounding box, and normalized by its largest extent; physical
size is provided separately in the state vector.
Both actor and critic receive the same 1,008-dimensional concatenation.
They have separate ReLU MLPs with hidden widths $(512,256,128)$ and
share the trainable shape encoder. The actor defines an 18-dimensional
Gaussian followed by $\tanh$; the critic predicts a scalar value.
Grasp-template labels and object IDs are not policy inputs.

\paragraph{Reward and optimization.}
Let $h_t$ be the object's center-of-mass elevation relative to the
episode start. For a nonfailure transition, the reward is
\begin{equation}
    r_t=\operatorname{clip}\!\left(\frac{h_t}{0.20\,\mathrm{m}},0,1\right)
        +\mathbf{1}_{\mathrm{success},t}\sum_{j=1}^{N_t}\gamma^j,
    \label{eq:method_reward}
\end{equation}
where $\gamma=0.99$ and $N_t$ is the number of control steps remaining
after the current step. Success requires at least 18\,cm elevation
maintained continuously for strictly more than 2\,s, checked at every
physics substep. A successful episode terminates immediately; the second
term credits the remaining horizon as a unit-reward absorbing state,
avoiding loss of future reward solely because success ends the episode.
A failure transition receives zero reward. We use no additional
pose-matching, contact, or grasp-diversity reward.

PPO~\cite{schulman2017proximalpolicyoptimizationalgorithms} jointly optimizes the teacher and shape encoder.
The changing reset distribution supplies learning opportunities near
different successful configurations and progressively earlier states.
The critic estimates their expected discounted returns, and the actor
learns feedback actions from PPO advantages. This implements the
value-function motivation through policy learning; it does not explicitly
differentiate the critic with respect to state to choose actions.
Diversity is expressed by the policy's behaviors across interaction
states, without an explicit multimodal action head or a diversity loss.

\subsection{Distillation to Deployable Observations}
\label{sec:method_student}

The student removes simulated tactile, contact, and dynamics information.
Its 142-dimensional state contains current and previous object-to-hand
poses, the current hand-to-base pose, current and previous finger angles,
finger control targets, seven mesh-geometry descriptors, and the four
most recent normalized actions. Poses use translation and a six-dimensional
rotation representation. A copied, frozen teacher shape encoder supplies
256 additional features. The resulting 398-dimensional input feeds a
ReLU MLP with widths $(512,256,128)$ and an 18-dimensional $\tanh$ output.
Real-world execution uses calibrated motion-capture object poses,
robot feedback, and known object meshes; it requires no simulated
velocities, contact forces, or physical-property parameters.

We first perform behavior cloning (BC) on teacher-controlled rollouts,
then use a DAgger-style procedure~\cite{ross2011reductionimitationlearningstructured} with equal numbers
of teacher-controlled and student-controlled environments.
All distillation rollouts use 3B initialization.
In both groups, the fixed teacher labels the actual visited state with
$a_t^T=\tanh(\mu_T(o_t^T))$. The student minimizes normalized action MSE:
\begin{equation}
    \mathcal L_{\mathrm{imit}}
    =\mathbb E_{\mathcal B}
      \left[\frac{1}{18}\left\|\pi_S(o_t^S)-a_t^T\right\|_2^2\right].
    \label{eq:student_imitation}
\end{equation}
Here $\mathcal B$ is a finite rolling buffer; DAgger minibatches contain
equal amounts of data from the two controllers. The student remains in
control throughout its own episodes while the teacher supplies labels.
Action histories record the commands actually sent by that episode's
controller. Normal failed trajectories are retained, allowing supervision
on the states encountered during unsuccessful interactions and recovery.
Student optimization is supervised and does not use PPO.

\begingroup
\setlength{\parskip}{5pt plus 1pt minus 1pt}
\setlength{\intextsep}{8pt plus 2pt minus 2pt}
\setlength{\textfloatsep}{12pt plus 2pt minus 2pt}
\setlength{\floatsep}{10pt plus 2pt minus 2pt}
\section{Experiments}
\label{sec:experiments}

We evaluate Dr.~Grasp-RL along three dimensions: grasping success across
object geometries, diversity of successful grasp configurations, and
robustness through closed-loop adjustment and regrasping.

\begin{figure}[!htbp]
  \centering
  \begin{minipage}[t]{0.48\linewidth}
    \vspace{0pt}
\centering
    \includegraphics[width=\linewidth]{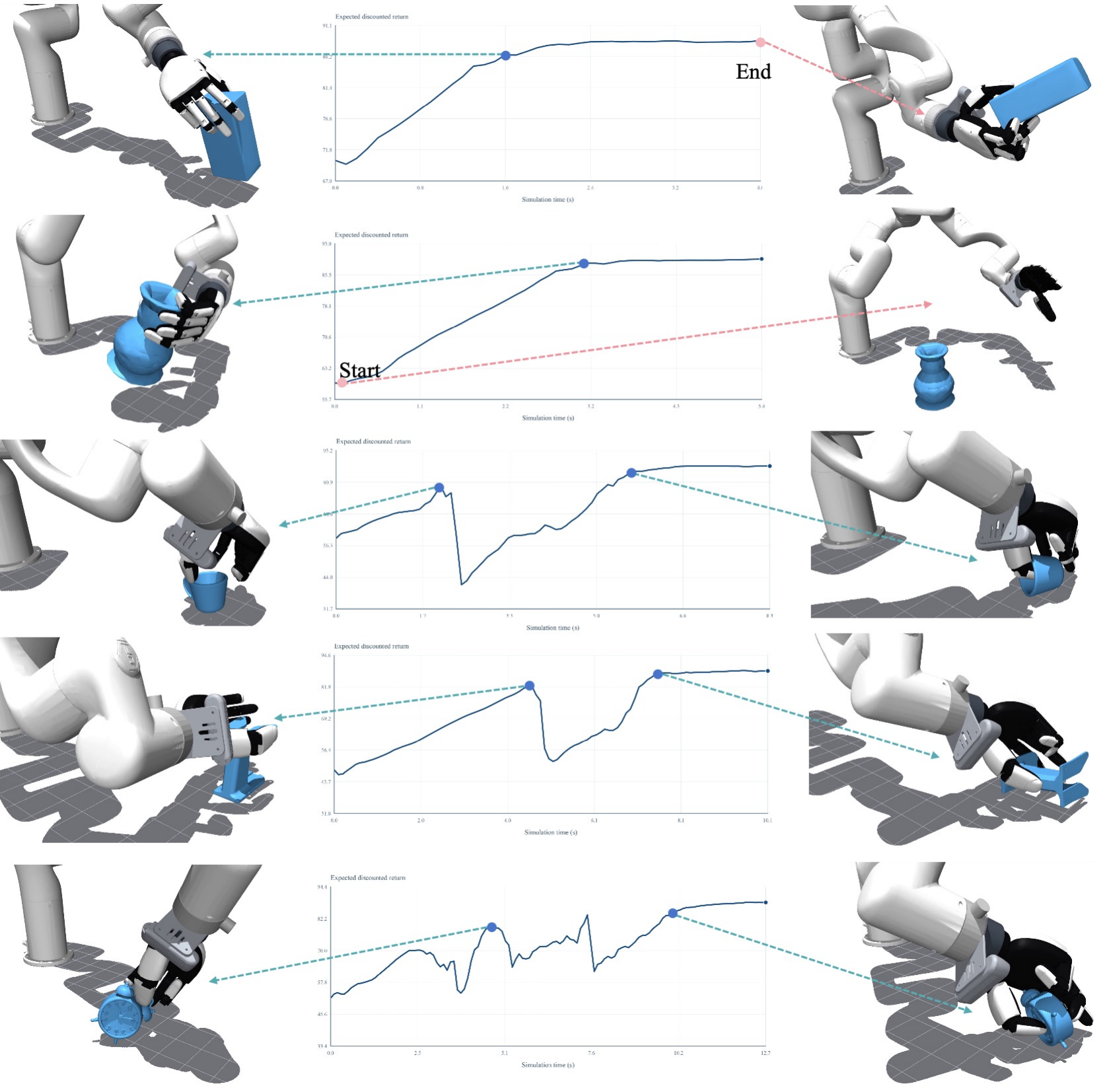}
    \caption{Value-function predictions along grasping trajectories.
    Snapshots illustrate the corresponding robot states.}
    \label{fig:value_function}
  \end{minipage}\hfill
  \begin{minipage}[t]{0.48\linewidth}
    \vspace{0pt}
\centering
    \captionof{table}{Grasping success rate (\%) on our shared
    training and test object assets. }
    \label{tab:success_rate_main}
    \begingroup
    \setlength{\tabcolsep}{4pt}
    \renewcommand{\arraystretch}{1.18}
    \scriptsize
    \resizebox{\linewidth}{!}{%
    \begin{tabular}{@{}lcccc@{}}
        \toprule
        & \multicolumn{2}{c}{With robot arm}
        & \multicolumn{2}{c}{Without robot arm} \\
        \cmidrule(lr){2-3} \cmidrule(lr){4-5}
        Method
        & Train.
        & Test.
        & Train.
        & Test. \\
        \midrule
        \multicolumn{5}{@{}l}{\textit{Generative grasp proposals with an execution controller}} \\
        Dexonomy~\cite{chen2025dexonomysynthesizingdexterousgrasp}
        & 58.16 & 46.32 & 65.81 & 53.45 \\
        DexGrasp Anything~\cite{zhong2025dexgraspanythinguniversalrobotic} (DGN)
        & 52.34 & 40.06 & 64.59 & 49.15 \\
        DexGrasp Anything~\cite{zhong2025dexgraspanythinguniversalrobotic} (UDG)
        & 50.10 & 37.97 & 62.46 & 49.35 \\
        \midrule
        \multicolumn{5}{@{}l}{\textit{RL-based closed-loop grasping and policy distillation}} \\
        UniDexGrasp++~\cite{wan2023unidexgraspimprovingdexterousgrasping}
        & 40.71 & 14.12 & 55.02 & 43.21 \\
        UniGraspTransformer~\cite{wang2025unigrasptransformersimplifiedpolicydistillation}
        & 56.46 & 40.51 & 80.38 & 65.14 \\
        RobustDexGrasp~\cite{zhang2025robustdexgrasprobustdexterousgrasping}
        & 80.38 & 61.80 & 74.64 & 57.96 \\
        \midrule
        \multicolumn{5}{@{}l}{\textit{Hybrid and other approaches}} \\
        UniDexGrasp~\cite{xu2023unidexgraspuniversalroboticdexterous}
        & 28.71 & 19.80 & 46.17 & 35.76 \\
        DemoGrasp~\cite{yuan2025demograspuniversaldexterousgrasping} (fixed init state)
        & 85.69 & 80.89 & 80.69 & 73.51 \\
        DemoFunGrasp~\cite{mao2025universaldexterousfunctionalgrasping} (fixed init state)
        & 65.53 & 57.24 & 50.93 & 38.99 \\
        DemoGrasp~\cite{yuan2025demograspuniversaldexterousgrasping}
        & 64.69 & 56.96 & 41.00 & 35.00 \\
        DemoFunGrasp~\cite{mao2025universaldexterousfunctionalgrasping}
        & 45.86 & 38.85 & 49.43 & 38.64 \\
        \midrule
        Ours (low quality stage1)
        & 66.89 & 54.18 & 76.67 & 62.83 \\
        Ours (w/o seed poses)
        & 11.10 & 9.68 & 10.81 & 9.53 \\
        Ours (teacher)
        & 97.44 & 92.11 & 98.92 & 93.23 \\
        \textbf{Ours (student)}
        & \textbf{95.74} & \textbf{88.06} & \textbf{97.82} & \textbf{91.87} \\
        \bottomrule
    \end{tabular}%
    }
    \endgroup
  \end{minipage}
\end{figure}

\subsection{Grasping Success Rate}\label{sec:success_rate}

\paragraph{Evaluation setup.}
Table~\ref{tab:success_rate_main} compares methods on the final
418-object training set and the 5,031-object test partition.
Our teacher and student are evaluated from ten different initial states
per object, with success defined as a COM rise of at least 18\,cm
maintained for more than 2\,s within a 20\,s episode.
Arm-mounted evaluation additionally accounts for IK and forbidden robot
collisions; hand-only evaluation removes the physical arm.
Baselines retain their configured native or adapted embodiments,
initialization, and termination rules, so the comparison concerns
complete grasping systems on shared object assets.
\begin{table}[!htbp]
    \centering
    \caption{Grasping success rate (\%) by object asset dataset.
Each column uses identical object IDs, scales, and asset preprocessing
for all methods. The evaluated datasets contain over 20,000 object
assets in total.}
    \label{tab:success_rate_by_asset_dataset}
    \begingroup
    \setlength{\tabcolsep}{3pt}
    \renewcommand{\arraystretch}{1.15}
    \scriptsize
    \resizebox{.8\linewidth}{!}{%
    \begin{tabular}{@{}lcccccc@{}}
        \toprule
        \shortstack{Method}
        & \shortstack{DexGraspNet\\(5000+)}
        & \shortstack{Objaverse\\(3500+)}
        & \shortstack{YCB\\(72)} 
        & \shortstack{DGA\\(9000+)}
        & \shortstack{EGAD\\(2330)}
        & \shortstack{VisualDexterity\\(150)} \\
        \midrule
        \multicolumn{7}{@{}l}{\textit{With robot arm}} \\
        \addlinespace[2pt]
        RobustDexGrasp~\cite{zhang2025robustdexgrasprobustdexterousgrasping} & 63.23 & 56.89 & 41.10 & 51.88 & 89.23 & 63.33\\
        DemoGrasp~\cite{yuan2025demograspuniversaldexterousgrasping} (fixed init state)  & 81.26 & 84.39 & 57.53 & 79.80 & 87.55 & \textbf{97.33}\\
        Ours (teacher) & 92.52 & 93.09 & 83.01 & 88.21 & 96.18 & 96.87\\
        \textbf{Ours (student)} & \textbf{88.65} & \textbf{88.42} & \textbf{70.27} & \textbf{81.74} & \textbf{94.69} & 94.67\\
        \midrule
        \multicolumn{7}{@{}l}{\textit{Without robot arm}} \\
        \addlinespace[2pt]
        UniGraspTransformer~\cite{wang2025unigrasptransformersimplifiedpolicydistillation} & 66.32 & 78.60 & 68.49 & 75.27 & 70.47 & 97.33\\
        DemoGrasp~\cite{yuan2025demograspuniversaldexterousgrasping} (fixed init state) & 74.06 & 74.76 & 61.64 & 72.97 & 52.96 & 88.00\\
        Ours (teacher) & 93.67 & 95.56 & 89.04 & 90.66 & 97.67 & 98.67\\
        \textbf{Ours (student)} & \textbf{92.33} & \textbf{94.91} & \textbf{89.18} & \textbf{89.06} & \textbf{97.45} & \textbf{98.87}\\
        \bottomrule
    \end{tabular}%
    }
    \endgroup
\end{table}
\vspace{20pt}

\paragraph{Ablation on seed initialization.}
We also evaluate two variants in Table~\ref{tab:success_rate_main}.
\textit{Ours (low quality stage1)} uses lower-quality seed poses
generated by Lightning-Grasp~\cite{yin2025lightninggrasphighperformance}. Its reduced final performance suggests
that weak grasp acquisition in Stage 1 limits subsequent learning,
with Stages 2 and 3 unable to fully compensate.
\textit{Ours (w/o seed poses)} removes near-grasp initialization
and instead uses dense geometric rewards analogous to the
optimization energies used for grasp synthesis.
Despite using the same number of training steps as the full method,
this variant achieves substantially lower success,
suggesting that reward shaping alone does not substitute for
learning from diverse, physically feasible near-grasp states.
Together, these comparisons highlight the importance of both
seed quality and initialization strategy.

\paragraph{Success and cross-dataset transfer.}
On the test partition, the teacher achieves 92.11\% success with an arm
and 93.23\% without an arm; the student achieves 88.06\% and 91.87\%,
respectively. Both exceed all baselines listed in
Table~\ref{tab:success_rate_main}, while the student's performance
shows that much of the teacher's capability transfers to deployable inputs.
Table~\ref{tab:success_rate_by_asset_dataset} further evaluates transfer
without fine-tuning across DexGraspNet, Objaverse, YCB, DGA, EGAD, and
VisualDexterity. Both policies outperform the listed baselines on five
of the six arm-mounted asset sets and all six hand-only sets.
These results support transfer across object geometries while preserving
strong performance under the constraints of an attached arm.

\subsection{Grasp Diversity}
\label{sec:grasp_diversity}
\begin{figure}[!htbp]
    \centering
    \includegraphics[width=.7\linewidth,height=.72\textheight,keepaspectratio]{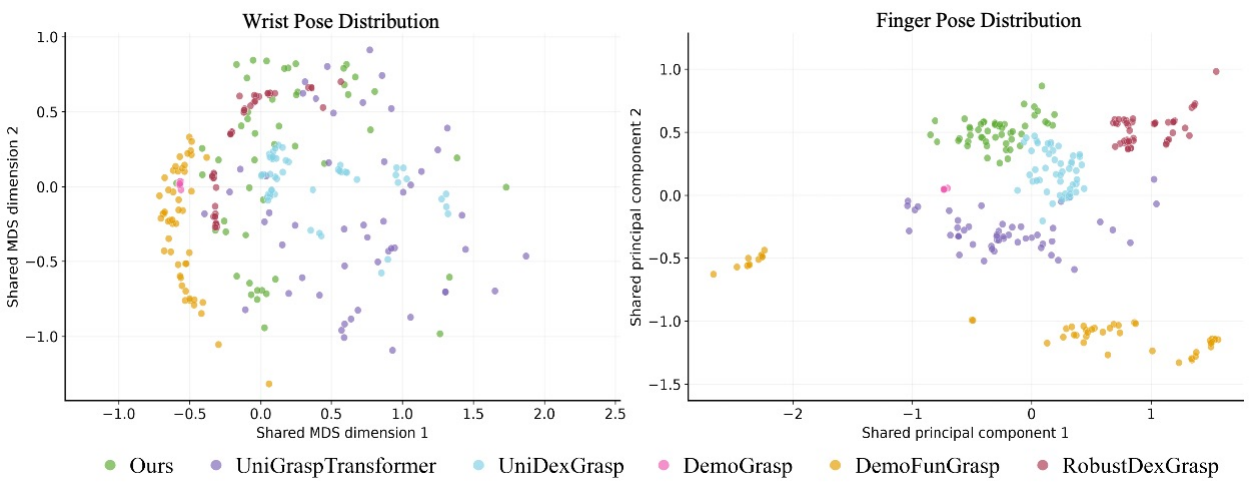}
    \caption{Distributions of wrist poses and finger configurations.}
    \label{fig:pose_distribution}
\end{figure}
\paragraph{Distinct high-value grasping modes.}
Figure~\ref{fig:diverse_grasps} shows multiple successful grasping
modes across training and unseen objects, with variations in wrist
orientation, contact side, and finger configuration.
The final row further illustrates diverse ways of approaching
the same object.
The teacher critic assigns consistently higher $V_0$ values to
the displayed grasp states than to the approach states;
along the illustrated approaches, values tend to increase as
the hand moves closer to the object.
Here $V_0$ is computed by setting only the critic's episode-progress
input to zero, leaving all other inputs unchanged.
Distinct successful configurations with similarly high values
are consistent with a multi-peak interpretation of the learned
value landscape.

Figure~\ref{fig:value_function} complements these snapshots with
student rollouts: predicted values generally increase toward
successful lifting, but can fall after unsuccessful interactions
and recover as the policy adjusts its grasp. Together, these observations suggest that the multiple high-value
grasping modes in Figure~\ref{fig:diverse_grasps} support recovery
through alternative grasps, as illustrated by the last three
rollouts in Figure~\ref{fig:value_function}, highlighting how
behavioral diversity enables dexterous adaptation.

\noindent
\begin{minipage}[t]{0.48\linewidth}
  \vspace{0pt}
  \setlength{\parskip}{5pt plus 1pt minus 1pt}
  \paragraph{Separating wrist placement and finger configuration.}
In Figure~\ref{fig:pose_distribution}, we compare 50 successful grasps distribution per method on a fixed cuboid
with unchanged object pose.
Repeated grasp outcomes are retained.
Each grasp is read at the first recorded frame reaching 1\,cm
COM elevation within the final contact episode that leads to
task success.
Wrist pose consists of object-relative translation and a unit
quaternion.
Finger configuration comprises five fingertip positions in a
calibrated wrist-aligned palm frame, normalized by the hand's
palm width. We uses shared MDS for wrist poses,
combining normalized translation and rotation-geodesic distances,
and shared PCA for the 15-dimensional fingertip descriptors.

In the shared wrist-pose projection, our green distribution covers a
broad region, comparable in spread to UniGraspTransformer and broader than the remaining
methods shown. UniGraspTransformer obtains diversity through the
distillation of separately trained object-specific experts; our
student inherits it from a single generalist teacher trained end to end
across objects with MLP-based PPO.
The coverage contrast is more pronounced for wrist pose than for
finger configuration, whose feasible patterns are highly constrained by
the object's surface.
\end{minipage}\hfill
\begin{minipage}[t]{0.46\linewidth}
  \vspace{0pt}
\centering
    \captionsetup{type=figure}
    \includegraphics[width=\linewidth]{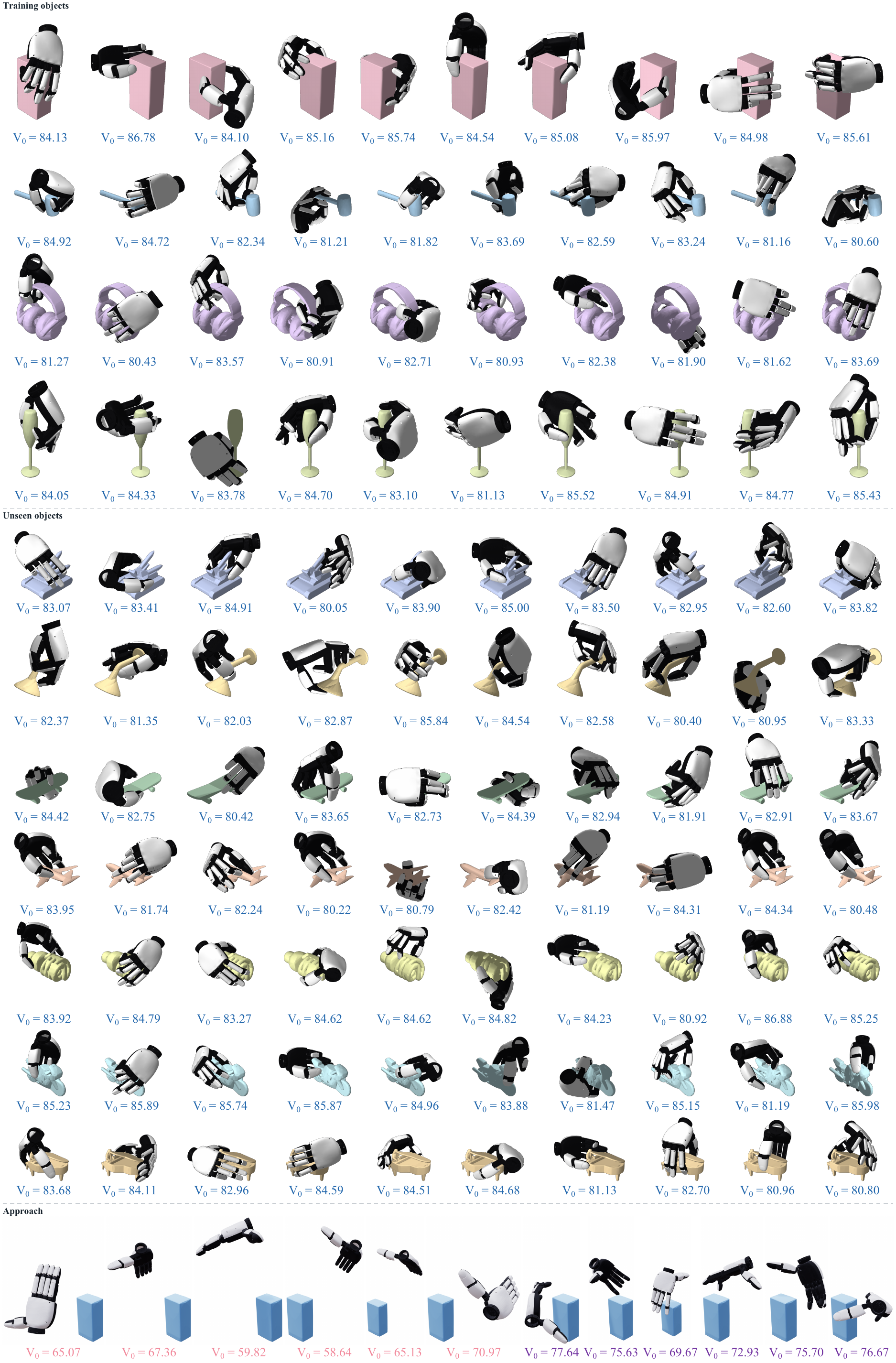}
    \caption{Diverse grasping and approach behaviors.}
    \label{fig:diverse_grasps}
\end{minipage}
\par\medskip

\subsection{Grasp Robustness}
\label{sec:grasp_robustness}
\label{sec:regrasping}

\begin{table}[!htbp]
    \centering
    \caption{Closed-loop regrasping within a 20-second episode on our object
    assets. SR is the overall episode success rate (\%).
    $p_1$, $p_2$, $p_3$, and $p_{\geq4}$ are the percentages of all successful
    episodes whose first successful grasp occurs on attempt 1, 2, 3, or
    $\geq4$, respectively. }
    \label{tab:regrasp_success_breakdown}
    \begingroup
    \setlength{\tabcolsep}{3pt}
    \renewcommand{\arraystretch}{1.12}
    \scriptsize
    \resizebox{.8\linewidth}{!}{%
    \begin{tabular}{@{}lcccccccccc@{}}
        \toprule
        & \multicolumn{5}{c}{Our trained object assets}
        & \multicolumn{5}{c}{Our test object assets} \\
        \cmidrule(lr){2-6} \cmidrule(lr){7-11}
        Method
        & \shortstack{Overall\\SR $\uparrow$}
        & \shortstack{1st\\$p_1$}
        & \shortstack{2nd\\$p_2$}
        & \shortstack{3rd\\$p_3$}
        & \shortstack{$\geq$4th\\$p_{\geq4}$}
        & \shortstack{Overall\\SR $\uparrow$}
        & \shortstack{1st\\$p_1$}
        & \shortstack{2nd\\$p_2$}
        & \shortstack{3rd\\$p_3$}
        & \shortstack{$\geq$4th\\$p_{\geq4}$} \\
        \midrule
        \multicolumn{11}{@{}l}{\textit{With robot arm}} \\
        UniDexGrasp++~\cite{wan2023unidexgraspimprovingdexterousgrasping} & 40.71 & 1.000 & 0 & 0 & 0 & 14.12 & 1.000 & 0 & 0 & 0 \\
        UniGraspTransformer~\cite{wang2025unigrasptransformersimplifiedpolicydistillation} (state-based) & 56.46 & 0.996 & 0.004 & 0 & 0 & 40.51 & 0.997 & 0.002 & 0.001 & 0 \\
        RobustDexGrasp~\cite{zhang2025robustdexgrasprobustdexterousgrasping} & 80.38 & 0.994 & 0.006 & 0 & 0 & 61.80 & 0.992 & 0.008 & 0 & 0 \\
        Ours (teacher) & 97.44 & 0.933 & 0.053 & 0.011 & 0.003 & 92.11 & 0.859 & 0.083 & 0.027 & 0.031 \\
        \textbf{Ours (student)} & \textbf{95.74} & 0.899 & 0.068 & 0.022 & 0.010 & \textbf{88.06} & 0.806 & 0.106 & 0.043 & 0.045 \\
        \midrule
        \multicolumn{11}{@{}l}{\textit{Without robot arm}} \\
        UniDexGrasp++~\cite{wan2023unidexgraspimprovingdexterousgrasping} & 55.02 & 0.952 & 0.030 & 0.009 & 0.009 & 43.21 & 0.937 & 0.052 & 0.006 & 0.004 \\
        UniGraspTransformer~\cite{wang2025unigrasptransformersimplifiedpolicydistillation} (state-based) & 80.38 & 0.917 & 0.077 & 0.006 & 0 & 65.14 & 0.932 & 0.060 & 0.006 & 0.002 \\
        RobustDexGrasp~\cite{zhang2025robustdexgrasprobustdexterousgrasping} & 74.64 & 0.984 & 0.013 & 0.003 & 0 & 57.96 & 0.984 & 0.016 & 0 & 0 \\
        Ours (teacher) & 98.92 & 0.962 & 0.030 & 0.006 & 0.002 & 93.23 & 0.877 & 0.066 & 0.024 & 0.033 \\
        \textbf{Ours (student)} & \textbf{97.82} & 0.942 & 0.045 & 0.010 & 0.003 & \textbf{91.87} & 0.855 & 0.078 & 0.033 & 0.034 \\
        \bottomrule
    \end{tabular}%
    }
    \endgroup
\end{table}

\paragraph{Closed-loop regrasping.}
Table~\ref{tab:regrasp_success_breakdown} reports overall success and
the attempt on which each successful episode first succeeds.
An attempt begins with finger--object contact; losing all such contact
for 0.3\,s ends the attempt, and subsequent contact starts another.
Palm or arm contact alone does not count.
Our policy continues from the resulting state after a nonterminal
failed grasp, without resetting the scene or policy history.
Attempt shares are conditioned on eventual success and must therefore
be interpreted together with overall success rate.

On the test partition with an arm, 14.1\% of teacher successes and
19.4\% of student successes occur after the first attempt.
Together with their overall SRs of 92.11\% and 88.06\%, these recovered
episodes contribute approximately 13.0 and 17.1 percentage points to
total success. The corresponding shares are larger than on the training
objects, indicating that retries remain useful under shape variation.
In comparison, at most 0.8\% of successes occur after the first attempt
for the arm-mounted baselines listed in the table.
Recovery also persists without the arm: approximately 12.3\% and
14.5\% of teacher and student successes, respectively, require retries.
Thus, high episode success is accompanied by successful recovery from
intermediate failures, rather than only immediate grasp acquisition.

\begin{figure}[!htbp]
    \centering
    \includegraphics[width=.97\textwidth,height=.72\textheight,keepaspectratio]{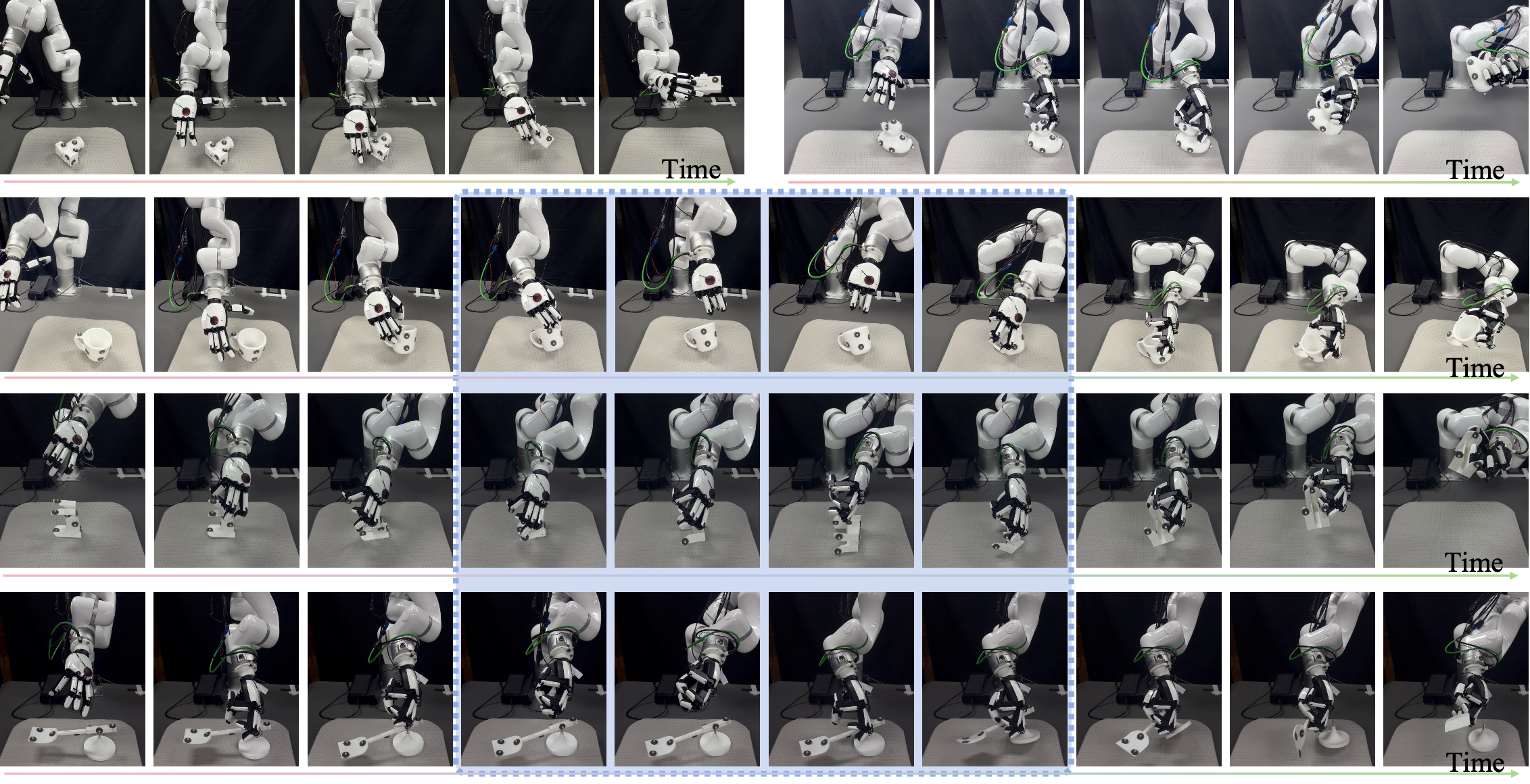}
    \caption{Real-world grasping sequences.
    Frames are ordered chronologically from left to right.
    Highlighted regions illustrate grasp adjustments during execution.}
    \label{fig:real_quality}
    
\end{figure}

\paragraph{Real-world adaptation.}
Figure~\ref{fig:real_quality} shows the student deployed with motion-capture
object poses, known object meshes, and robot feedback.
The first row illustrates successful single-attempt grasps.
The remaining three rows show unsuccessful contacts followed by online
changes in wrist pose, finger configuration, or approach before lifting
the object successfully. These sequences demonstrate that the deployed
policy reacts to the evolving interaction instead of replaying a fixed
grasp trajectory.

\paragraph{Coupling diversity and robustness.}
A diverse grasping repertoire provides alternative actions when a
previous approach becomes unsuitable, while feedback allows the policy
to use these alternatives from the state left by a failed attempt.
The observed changes in grasp configuration during successful retries
are consistent with this connection: recovery can involve selecting
a different contact arrangement, rather than mechanically repeating
the first unsuccessful motion.
Together, the grasp distributions, regrasping statistics, and real-world
sequences support the value of learning diverse behaviors within a
single closed-loop policy.

\par
\endgroup
\section{Conclusion}
We presented Dr.~Grasp-RL, which uses diverse resets and a
time-reversed curriculum to train a dexterous grasping policy with PPO.
The policy acquires multiple grasping modes without a generative
architecture and combines this diversity with closed-loop control
for real-time adjustment and recovery.
Simulation and real-world experiments demonstrate strong
generalization across object geometries and robust execution.

\paragraph{Implications.}
These findings motivate a control-centered view of robot learning.
Research paradigms and narratives borrowed from other domains
may not be the best fit for the distinctive demands of robotics.
We therefore advocate developing models and architectures grounded
in robotics' own requirements for physical interaction.

\clearpage
\pdfbookmark[1]{References}{references}
\bibliographystyle{abbrvnat}
{\raggedright\bibliography{main}}

\clearpage
\appendix
\section*{Appendix}
\phantomsection
\addcontentsline{toc}{section}{Appendix}
\makeatletter
\setlength{\@fptop}{0pt}
\setlength{\@dblfptop}{0pt}
\setlength{\@dblfpsep}{18pt}
\makeatother
\setcounter{figure}{0}\setcounter{table}{0}\setcounter{equation}{0}
\renewcommand{\thefigure}{S\arabic{figure}}
\renewcommand{\thetable}{S\arabic{table}}
\renewcommand{\theequation}{S\arabic{equation}}
\renewcommand{\theHfigure}{appendix.\arabic{figure}}
\renewcommand{\theHtable}{appendix.\arabic{table}}
\renewcommand{\theHequation}{appendix.\arabic{equation}}

\begin{figure}[p]
    \centering
    \includegraphics[
        width=\textwidth,
        height=.9\textheight,
        keepaspectratio
    ]{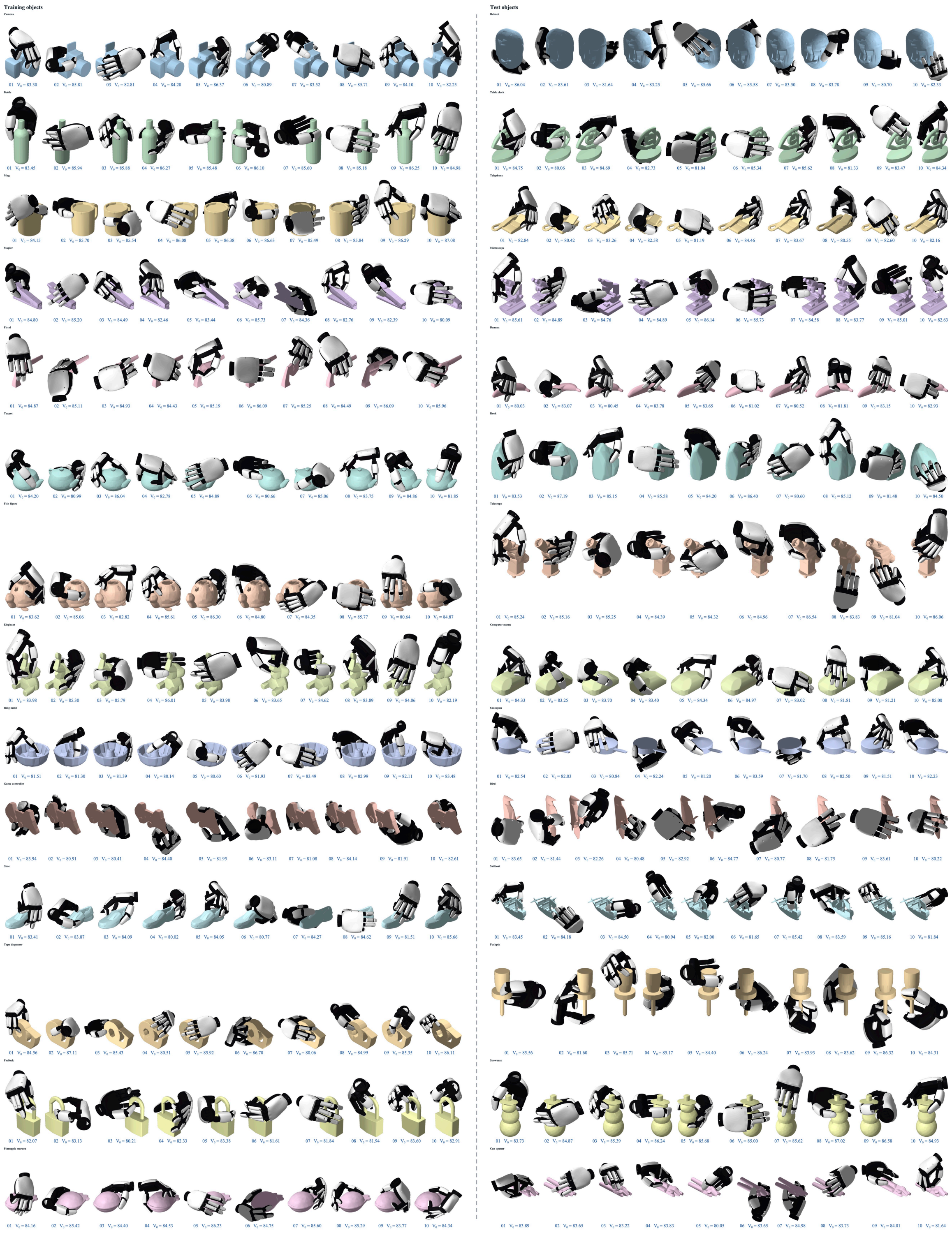}

    \caption{
    Qualitative grasp diversity on different training objects (left)
    and unseen objects (right), separated by the vertical
    dashed line. All displayed states belong to ultimately successful grasps
    and are captured at the first recorded lift of at least
    1\,cm. We hide the robot arm and
    ground in rendering for better clarity and views are fixed in the object frame.
 }
    \label{fig:appendix_multi_object_diversity}
\end{figure}

\section{Observation and Network Specifications}
\label{sec:supp_observations}

\paragraph{Coordinate conventions.}
Let $B$, $H$, and $O$ denote the robot base, hand root, and object mesh frames.
$T_{HO}$ expresses the object in the instantaneous hand-root frame;
$T_{BH}$ expresses the hand root in the robot-base frame.
The hand frame is the XHand root link, not the arm flange. Its pose is obtained
by forward kinematics with the fixed flange-to-hand mounting transform.
Every pose in the network input has nine components: three translation
coordinates and a six-dimensional rotation representation containing the
first two columns of the rotation matrix, concatenated column by column
$(R_{00},R_{10},R_{20},R_{01},R_{11},R_{21})$. Quaternions used in analysis and
visualization are not the policy's rotation input representation.

\paragraph{Normalization.}
Translations and mesh geometry are divided by $0.5$\,m. A joint angle is mapped
to $(q-q_c)/q_h$, where $q_c=(q_{\min}+q_{\max})/2$ and
$q_h=(q_{\max}-q_{\min})/2$ are obtained from the robot joint limits.
Angular velocities are divided by $5$\,rad/s, whereas linear velocities
remain in m/s. State vectors are clipped elementwise to $[-10,10]$ after
normalization. Tables~\ref{tab:supp_teacher_input} and
\ref{tab:supp_student_input} give the exact concatenation order, using
zero-based inclusive indices.

\begin{table}[!hbp]
\centering
\begin{minipage}{0.80\linewidth}
\centering
\captionsetup{font=footnotesize,skip=4pt}
\setlength{\parskip}{0pt}
\caption{Teacher input: 152 state scalars, 600 tactile scalars, and a 256-dimensional shape embedding, totaling 1,008 dimensions. Both actor and critic receive this same input.}
\label{tab:supp_teacher_input}
\footnotesize\setlength{\tabcolsep}{3pt}\renewcommand{\arraystretch}{1.0}
\setlength{\aboverulesep}{1.5pt}\setlength{\belowrulesep}{1.5pt}
\begin{tabular*}{\linewidth}{@{\extracolsep{\fill}}rr>{\raggedright\arraybackslash}p{.35\linewidth}>{\raggedright\arraybackslash}p{.42\linewidth}@{}}
\toprule
Indices & Dim. & Quantity & Representation / normalization \\
\midrule
0--8 & 9 & Object pose in $H$ & Translation / 0.5\,m; rotation 6D \\
9--17 & 9 & Hand-root pose in $B$ & Translation / 0.5\,m; rotation 6D \\
18--20 & 3 & Object COM position in $H$ & Position / 0.5\,m \\
21--26 & 6 & Object relative linear and angular velocity in $H$ & Linear velocity; angular velocity / 5\,rad/s \\
27--32 & 6 & Hand-root linear and angular velocity in $B$ & Linear velocity; angular velocity / 5\,rad/s \\
33--51 & 19 & Measured arm and hand joint angles & 7 arm + 12 hand; joint-limit normalization \\
52--70 & 19 & Measured arm and hand joint velocities & 7 arm + 12 hand; / 5\,rad/s \\
71--82 & 12 & Current finger position targets & Joint-limit normalization \\
83--100 & 18 & Previous issued policy action & Normalized action in $[-1,1]$ \\
101--106 & 6 & Gravity expressed in $H$ and $B$ & Two 3D vectors; / 10\,m/s$^2$ \\
107--112 & 6 & Support-plane normal in $H$ and $B$ & Two 3D unit vectors \\
113--114 & 2 & Hand-root and base heights above support plane & / 0.5\,m \\
115--121 & 7 & Mesh geometry in $O$ & AABB center (3), extents (3), maximum extent (1); / 0.5\,m \\
122--140 & 19 & Object and environment dynamics & Mass (1), COM offset (3), inertia (6), friction (6), gravity (3); see text \\
141--144 & 4 & Task progress and time & COM rise / 0.2\,m; lowest-point height / 0.1\,m; age / 20\,s; hold time / 2\,s \\
145--150 & 6 & Hand--object normal contact forces & Five fingers and palm; / 10\,N \\
151 & 1 & Object support force from the floor & / 10\,N \\
\midrule
152--751 & 600 & Distributed tactile readings & 120 touch sensors per finger; $\tanh(f/2\,\mathrm{N})$ \\
752--1007 & 256 & Learned mesh shape feature & PointNet-style encoding of 512 surface points \\
\bottomrule
\end{tabular*}
\end{minipage}
\end{table}

\paragraph{Privileged dynamics and contact inputs.}
The 19-dimensional dynamics block contains mass in kg; object-frame COM
offset divided by $0.5$\,m; the six distinct inertia entries
$(I_{xx},I_{yy},I_{zz},I_{xy},I_{xz},I_{yz})$ divided by
$10^{-3}$\,kg\,m$^2$; three effective hand--object friction coefficients
(sliding, torsional, and rolling); three effective floor--object friction
coefficients; and world gravity divided by $10$\,m/s$^2$.
Tactile readings are averaged over the physics substeps of a control interval
before the bounded transformation. The separate six-dimensional force block
aggregates normal contact forces over the five fingers and palm.

\paragraph{Shape encoding.}
For each object, 512 points are sampled from its mesh surface in $O$.
The point set is centered at the axis-aligned bounding-box center and divided
by its maximum extent. A shared pointwise MLP has widths $3$--$64$--$128$--$256$,
with ReLU after the first two layers. Concatenated max and mean pooling give
512 values, followed by a linear $512$--$256$ projection. Explicit geometry
metadata preserves scale information removed by point normalization.
The teacher jointly trains this encoder with its actor and critic. The student
copies and freezes the final teacher encoder; the shape feature can therefore
be cached once per known mesh at deployment.

\paragraph{Teacher architecture.}
Separate actor and critic MLPs have hidden widths $512$, $256$, and $128$,
with ReLU activations. The actor produces 18 Gaussian means and has a learned,
state-independent 18-dimensional log standard deviation. Actions are sampled
and passed through $\tanh$, with the corresponding log-probability correction.
Deterministic evaluation uses $\tanh$ of the mean. The critic produces one
scalar. Including the shape encoder and log standard deviation, the teacher
has 1,537,061 trainable parameters. There is no recurrent module or explicit
grasp-mode, object-identity, or reset-identity input.

\begin{table}[!htbp]
\centering
\begin{minipage}{0.80\linewidth}
\centering
\captionsetup{font=footnotesize,skip=4pt}
\setlength{\parskip}{0pt}
\caption{Final student input: 142 state scalars plus the frozen 256-dimensional shape embedding, totaling 398 dimensions. The student directly outputs 18 bounded action components.}
\label{tab:supp_student_input}
\footnotesize\setlength{\tabcolsep}{3pt}\renewcommand{\arraystretch}{1.0}
\setlength{\aboverulesep}{1.5pt}\setlength{\belowrulesep}{1.5pt}
\begin{tabular*}{\linewidth}{@{\extracolsep{\fill}}rr>{\raggedright\arraybackslash}p{.35\linewidth}>{\raggedright\arraybackslash}p{.42\linewidth}@{}}
\toprule
Indices & Dim. & Quantity & Source / convention \\
\midrule
0--8 & 9 & Current object pose in $H$ & Mocap object pose and hand forward kinematics \\
9--17 & 9 & Previous object pose in $H$ & Previous frame's representation, as stored at that time \\
18--26 & 9 & Current hand-root pose in $B$ & Robot forward kinematics \\
27--38 & 12 & Current measured finger angles & Joint-limit normalization \\
39--50 & 12 & Previous measured finger angles & Joint-limit normalization \\
51--62 & 12 & Current commanded finger targets & Joint-limit normalization \\
63--69 & 7 & Mesh geometry in $O$ & Same center, extents, and maximum extent as the teacher \\
70--141 & 72 & Four previous issued actions & $a_{t-1},a_{t-2},a_{t-3},a_{t-4}$, newest first \\
\midrule
142--397 & 256 & Mesh shape feature & Frozen teacher shape encoder \\
\bottomrule
\end{tabular*}
\end{minipage}
\end{table}

\paragraph{Student architecture and history.}
The student actor is a $398$--$512$--$256$--$128$--$18$ MLP, with ReLU hidden
layers and a final $\tanh$. It has 370,834 trainable parameters plus 172,928
frozen shape-encoder parameters. The action history stores commands actually issued to each world. It
contains neither measured pose changes nor the teacher labels assigned to
student-controlled worlds. On reset, the previous object pose
and previous finger angles equal their current values; all four action slots
are zero. Each environment maintains its own history. The previous object
pose is not re-expressed in the new hand frame.
The student excludes arm joint angles as direct inputs, all joint velocities,
relative velocities, tactile/contact measurements, object dynamics, support
plane descriptors, gravity, episode age, lift height, and hold time.
Arm joint angles are still needed outside the network for forward kinematics.

\section{Control, Reward, and PPO}
\label{sec:supp_control}
\paragraph{Simulation and timing.}
The final teacher is trained with the xArm7 and 12-actuated-joint XHand in
mjlab-based GPU simulation. Physics runs at 1,000\,Hz and the policy at
10\,Hz, giving 100 physics substeps per action. Each episode lasts at most
200 policy steps (20\,s). The curriculum
randomizes reset states and, in Stage~3, object positions.

\begin{table}[!htbp]
\centering
\begin{minipage}{0.70\linewidth}
\centering
\captionsetup{font=footnotesize,skip=4pt}
\setlength{\parskip}{0pt}
\caption{Action vector and scaling. All components lie in $[-1,1]$ before scaling.}
\label{tab:supp_action}
\footnotesize\setlength{\tabcolsep}{3pt}\renewcommand{\arraystretch}{1.0}
\setlength{\aboverulesep}{1.5pt}\setlength{\belowrulesep}{1.5pt}
\begin{tabular*}{\linewidth}{@{\extracolsep{\fill}}rcl@{}}
\toprule
Indices & Dimension & Physical increment \\
\midrule
0--2 & 3 & Local $x,y,z$: 0.01\,m per component \\
3--5 & 3 & Local roll, pitch, yaw: $10^\circ$ each \\
6--17 & 12 & Finger angles: 0.2\,rad each \\
\bottomrule
\end{tabular*}
\end{minipage}
\end{table}

\paragraph{Hand-root actions and IK.}
Let $R_{BH}$ and $p_{BH}$ be the current hand pose and let
$(\Delta p,\Delta\phi,\Delta\theta,\Delta\psi)$ denote the scaled translation,
roll, pitch, and yaw increments. The desired pose is
\begin{align}
p^*_{BH}&=p_{BH}+R_{BH}\Delta p,\\
R^*_{BH}&=R_{BH}R_z(\Delta\psi)R_y(\Delta\theta)R_x(\Delta\phi).
\end{align}
A damped least-squares IK solver maps this target to the arm joints. The
solver uses up to 16 iterations, a rotational residual/Jacobian scale of
0.3, and a maximum joint update of 0.15\,rad per IK iteration. Its acceptance
tolerances are 1\,mm in position and 0.01\,rad in orientation. Accepted arm
joint targets are slew-limited to 1.5\,rad/s; an unsolved target retains the
previous command. Thus, specifying a Cartesian target does not teleport the
hand to that pose. The simulator executes joint position control throughout.

\paragraph{Finger commands.}
Finger increments are added to the measured finger positions and clipped to
joint limits. The resulting commanded targets are further slew-limited by
0.2\,rad per control step. The order is thumb bend and two thumb rotation
joints; index bend and two index joints; two middle-finger joints; two
ring-finger joints; and two little-finger joints. This is the same action
contract for teacher and student.

\paragraph{Lift reward and episode completion.}
Let $h_t$ be the object COM rise from reset and let $T=200$ be the control-step
limit. Away from failure, the dense height reward is
\begin{equation}
r_t^{\mathrm{height}}=\operatorname{clip}(h_t/0.20,0,1).
\end{equation}
Success requires $h\geq0.18$\,m continuously for strictly more than 2\,s.
This is checked at physics frequency, with the hold counter reset whenever
the condition is lost. At a successful terminal step, an absorbing-success
tail $\sum_{j=1}^{N_{\rm rem}}\gamma^j$ is added, where $N_{\rm rem}$ is the
number of remaining control steps. Failure transitions receive zero reward.
There is no grasp-pose tracking, contact-shape matching, or explicit diversity
reward. Terminal success, failure, and the 20\,s timeout all stop value
bootstrapping; a nonterminal rollout-buffer boundary does not.

\paragraph{PPO optimization.}
We use generalized advantage estimation and a clipped probability-ratio
objective. With $\rho_t=\pi_\theta(a_t|s_t)/\pi_{\rm old}(a_t|s_t)$,
\begin{align}
L_{\pi}&=-\mathbb{E}\!\left[\min\!\left(\rho_t\hat A_t,
\operatorname{clip}(\rho_t,1-\epsilon,1+\epsilon)\hat A_t\right)\right],\\
L_V&=\tfrac12\mathbb{E}[(V_\theta(s_t)-\hat G_t)^2],\\
L&=L_{\pi}+0.5L_V-0.001\mathcal H(\pi_\theta).
\end{align}
The plotted value loss is $L_V$, including the factor $1/2$.
Advantages are normalized over the rollout; distributed training uses global
statistics and averages gradients across GPUs. The point encoder is
re-evaluated during PPO updates so both heads train it with current weights.
The learning rate is divided by 1.5 when the monitored KL exceeds 0.04 and
multiplied by 1.5 below 0.01, within the range in
Table~\ref{tab:supp_ppo}. A hard KL guard can stop an update early.

\begin{table}[!htbp]
\centering
\begin{minipage}{0.75\linewidth}
\centering
\captionsetup{font=footnotesize,skip=4pt}
\setlength{\parskip}{0pt}
\caption{PPO hyperparameters shared across teacher stages.}
\label{tab:supp_ppo}
\footnotesize\setlength{\tabcolsep}{3pt}\renewcommand{\arraystretch}{1.0}
\setlength{\aboverulesep}{1.5pt}\setlength{\belowrulesep}{1.5pt}
\begin{tabular*}{\linewidth}{@{\extracolsep{\fill}}ll@{}}
\toprule
Parameter & Value \\
\midrule
Discount $\gamma$ / GAE $\lambda$ & 0.99 / 0.95 \\
Rollout horizon & 32 control steps (3.2\,s) \\
PPO epochs / minibatches per epoch & 4 / 8 \\
Maximum optimizer steps per update & 32 \\
Ratio clip $\epsilon$ & 0.2 \\
Optimizer & Adam \\
Initial learning rate (Stage 1) & $3\times10^{-4}$ \\
Learning-rate bounds & $[10^{-5},3\times10^{-4}]$ \\
KL target / hard limit & 0.02 / 0.08 \\
Value-loss weight / entropy weight & 0.5 / 0.001 \\
Maximum gradient norm & 0.5 \\
Initial log standard deviation & $-0.8$ \\
Log standard deviation bounds & $[-5,1]$ \\
Checkpoint interval & 5 PPO updates \\
\bottomrule
\end{tabular*}
\end{minipage}
\end{table}

\paragraph{Meaning of the value visualization.}
The critic estimates expected discounted return under the current policy,
not success probability. Episode age is one teacher input because the task
has a finite horizon. Our $V_0$ diagnostic replaces only the normalized age
component (state index 143) by zero while preserving all other inputs. It is a time-input
ablation, not a newly calibrated probability or a reset of the physical
state. PPO learns the actor through advantages; it does not execute the
literal state-space gradient $\nabla_s V$. The multi-peak landscape in the
teaser is a conceptual illustration rather than a measured global critic
surface.

\section{Reset Banks and Training Curriculum}
\label{sec:supp_curriculum}
\paragraph{Stage 1: near-grasp learning.}
The recorded initial bank contains 9,036 seeds over 418 objects. One environment is assigned to each seed. The final
Stage~1 checkpoint is saved after 194 PPO updates. From Stage~2 onward, we
retain the 9,036 seeds and their 418 objects. 

\paragraph{Opening and moving away.}
The natural open-hand joint vector, in the order specified above, is
\begin{equation}
q_{\rm open}=(.08,-.25,.08,0,.08,.08,.08,.08,.08,.08,.08,.08)
\end{equation}
in radians. ``Fully open'' refers to this vector, not to all joint lower
limits. For 2A, the fingers interpolate halfway from the source seed toward
$q_{\rm open}$. The hand-root pose is retained whenever feasible; otherwise,
a local adjustment is made subject to collision and IK checks.
For 2B, the reference pose is first adjusted with fingers fully open, then
moved outward by 2--8\,cm. The reference adjustment budget is initially
6\,cm/$35^\circ$; only failed entries may expand to 8\,cm/$45^\circ$.
This prevents an invalid fully open reference from being mistaken for a
valid retreat start. Both banks retain all 9,036 source indices.

\paragraph{Far, independent resets.}
The 2C bank contains 200 feasible open-hand robot configurations, shared
across objects. Its 5--25\,cm separation refers to a projected geometric
clearance along the object-to-palm direction, not simply the distance
between frame origins. The bank is checked against all 418 training objects.
Each environment independently and uniformly samples a bank entry on reset.
During Stage~3, separate 200-entry banks are constructed for 3A and 3B and
checked against object displacement envelopes. They can be farther away
than the 2C bank; the 5--25\,cm interval is not imposed on every 3B pose.
Object and robot reset draws are independent, including across GPU ranks.

\paragraph{Feasibility checks.}
Opening/retreat candidates are checked for robot--object and moving
robot--floor clearance (1\,mm), prohibited self-intersection, joint limits,
and endpoint IK. Reset-bank IK validation uses 0.5\,mm position and
0.005\,rad orientation tolerances. These are initialization checks, not
successful-trajectory filtering. The banks are generated before learning;
subsequent failed trajectories remain part of training.

\begin{table}[!htbp]
\centering
\begin{minipage}{0.80\linewidth}
\centering
\captionsetup{font=footnotesize,skip=4pt}
\setlength{\parskip}{0pt}
\caption{Actual environment allocations along the final teacher-training lineage. Counts in multi-GPU rows are \emph{per GPU}; four GPUs are used. Within each group, environments sample resets independently.}
\label{tab:supp_curriculum}
\footnotesize\setlength{\tabcolsep}{3pt}\renewcommand{\arraystretch}{1.0}
\setlength{\aboverulesep}{1.5pt}\setlength{\belowrulesep}{1.5pt}
\begin{tabular*}{\linewidth}{@{\extracolsep{\fill}}l>{\raggedright\arraybackslash}p{.40\linewidth}rl@{}}
\toprule
Phase & Composition & Environments & GPUs \\
\midrule
Stage 1 & One environment for each of 9,036 seeds & 9,036 & 1 \\
Stage 2, level 1 & 9,036 seeds + 9,036 2A & 18,072 & 1 \\
Stage 2, level 2 & 9,036 2A + 9,036 2B & 18,072 & 1 \\
Stage 2, level 3 (initial) & 9,036 2B + $418\times20$ 2C & 17,396 & 1 \\
Stage 2, level 3 (continued) & 2,259 2B + $418\times10$ 2C & 6,439 & 4 \\
Stage 3, 3A warm start & $418\times20$ 3A & 8,360 & 4 \\
Stage 3, level 1 & $418\times3$ 2C + $418\times20$ 3A & 9,614 & 4 \\
Stage 3, level 2 & $418\times1$ 2C + $418\times5$ 3A + $418\times17$ 3B & 9,614 & 4 \\
\bottomrule
\end{tabular*}
\end{minipage}
\end{table}

\paragraph{Stage 2 gates and replay of easier resets.}
Level~1 mixes seeds and 2A; level~2 mixes 2A and 2B; level~3 mixes 2B and 2C.
Promotion requires at least 60\% success on the harder group (2A or 2B).
The gate gives each reset index equal weight, uses its last 20 episode
outcomes, and requires at least 10 episodes for every reset index. This
avoids fast-resetting cases dominating the transition statistic. The
observed transitions occur after updates 82 and 313. Training continues
from the single-GPU run on four GPUs, partitioning the 9,036 2B sources into
four equally sized disjoint subsets. Table~\ref{tab:supp_curriculum} gives
the changed 2C allocation. PPO weights and optimizer state are continued.

\paragraph{Stage 3 position generalization.}
3A samples object offsets independently along $x$ and $y$ within
$[-5,5]$\,cm. 3B uses $x\in[-5,25]$\,cm and $y\in[-15,15]$\,cm, with
positive $x$ pointing away from the arm and $y$ lateral. The object's height,
orientation, scale, and mass remain unchanged at reset. The 2C retention
group uses the original fixed object position.
The recorded run begins with 37 updates of 3A-only training, followed by
two updates of mixed level~1 and 97 updates of mixed level~2. Promotion to
level~2 requires 70\% recent 3A success, averaged equally over the 418
objects, with a last-100-episodes window per object and at least 10 episodes
for every object. The final teacher is the stop checkpoint after 136
Stage~3 updates (global Stage~2/3 update index 1,017).
The retention groups continue rehearsing easier settings while the harder
reset distribution is learned.

\section{Behavior Cloning and DAgger}
\label{sec:supp_distillation}
\paragraph{Shared setup.}
Both distillation phases use the final teacher and the 3B reset distribution
only. Each of four GPUs hosts $418\times20=8,360$ environments, totaling
33,440. The student has the observation contract in
Table~\ref{tab:supp_student_input}; only these deployable quantities enter the student. The teacher uses its
privileged observation only to provide action labels. At each visited state, the target is
the teacher's deterministic bounded action $\tanh(\mu_T(s))$.
We minimize the mean squared error over all 18 normalized action components.
The shape encoder is frozen, and only the student actor is optimized.

\begin{table}[!htbp]
\centering
\begin{minipage}{0.75\linewidth}
\centering
\captionsetup{font=footnotesize,skip=4pt}
\setlength{\parskip}{0pt}
\caption{Student training settings. Batch and buffer counts below are per GPU.}
\label{tab:supp_distillation}
\footnotesize\setlength{\tabcolsep}{3pt}\renewcommand{\arraystretch}{1.0}
\setlength{\aboverulesep}{1.5pt}\setlength{\belowrulesep}{1.5pt}
\begin{tabular*}{\linewidth}{@{\extracolsep{\fill}}l>{\raggedright\arraybackslash}p{.41\linewidth}@{}}
\toprule
Parameter & Value \\
\midrule
Environments / GPUs & 8,360 / 4 \\
Collection horizon & 64 control steps \\
Replay capacity & 256 transitions per world \\
Gradient steps per collection & 256 \\
Batch size & 8,192 (32,768 globally) \\
Optimizer / learning rate & AdamW / $3\times10^{-4}$ \\
Weight decay / gradient norm limit & $10^{-5}$ / 1.0 \\
Validation minibatches per update & 8 \\
BC collection rounds & 43 \\
DAgger collection rounds & 33 \\
Checkpoint interval & 5 rounds \\
\bottomrule
\end{tabular*}
\end{minipage}
\end{table}

\paragraph{BC on teacher-visited states.}
All BC environments execute the frozen teacher. For every object and GPU,
15 worlds are assigned to training and five to validation. Training and
validation therefore use separate worlds and trajectories, but the same
object identities. The rolling buffers contain up to 1,605,120 training
and 535,040 validation transitions per GPU. We collect 64 steps, perform
256 updates using uniform sampling with replacement, and continue collection.
We do not first collect an unbounded static dataset and train it to convergence.
The final BC checkpoint follows 43 rounds and 92,026,880 collected teacher
transitions across GPUs. Failed trajectories are retained; invalid numerical
transitions are removed. The expert success recorded during BC describes
the data-collection policy, not the autonomous student.

\paragraph{DAgger on teacher- and student-visited states.}
DAgger starts from the final BC student with a fresh optimizer. For each
object and GPU, 10 worlds execute the teacher and 10 execute the current
student. Within each group, eight worlds are used for training and two for
validation. Separate rolling buffers are maintained for the two sources;
each source has 856,064 training and 214,016 validation slots per GPU.
Each training minibatch takes 4,096 samples from each source. The teacher
labels the student's actual visited states, including unsuccessful approaches
and recoveries; it does not take over control of these worlds.

The student is held fixed during each 64-step collection, then updated for
256 gradient steps. The next collection uses the updated student. Episodes
can span collection rounds, and histories record the actions actually
executed in each world. The final checkpoint follows 33 DAgger rounds and
70,625,280 additional collected transitions across both sources. The BC and
DAgger losses use the same action units and ordering, so their fitting errors
are comparable, while the state distributions differ.

\section{Learning Curves and Their Interpretation}
\label{sec:supp_curves}
\begin{figure}[ht]
\centering
\includegraphics[width=\textwidth,height=.72\textheight,keepaspectratio]{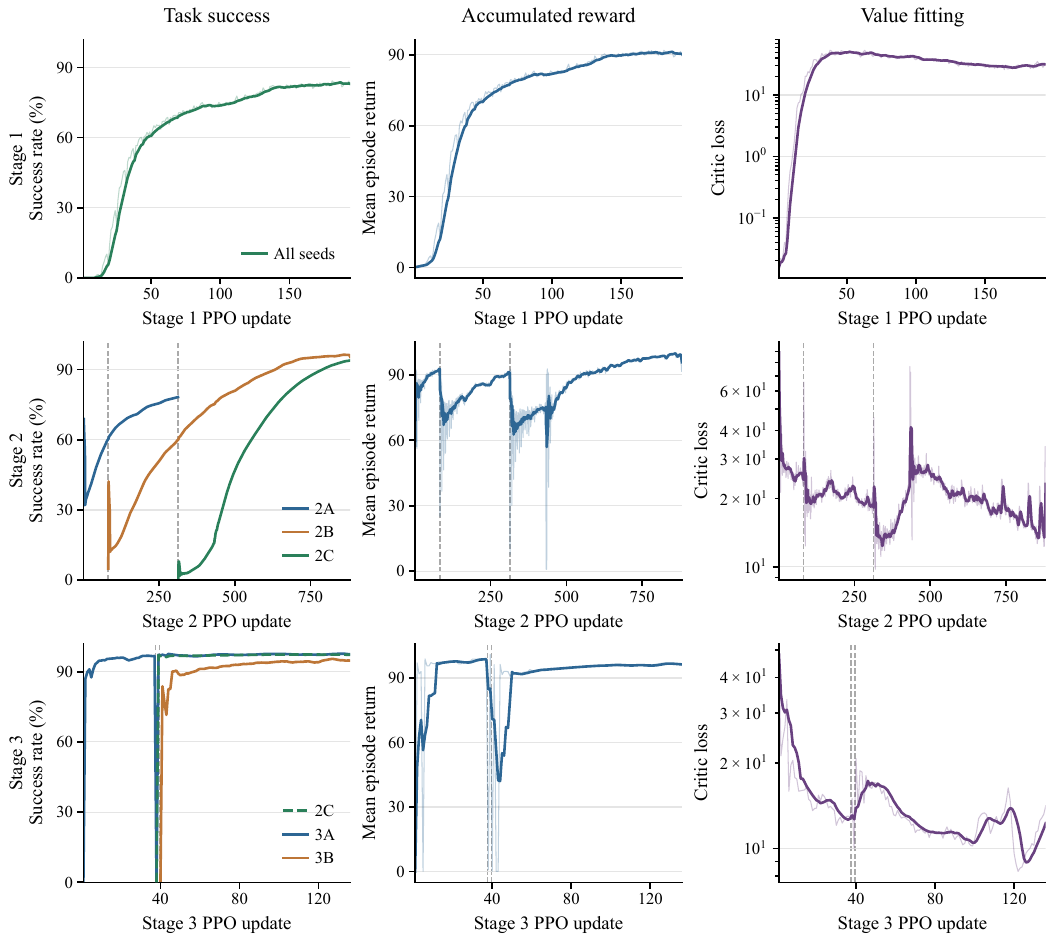}
\caption{Teacher learning through the three curriculum stages. Success is separated by active reset group in Stages~2 and 3. Return and value loss are computed over the current training mixture. Vertical dashed lines indicate level changes, including the 3A-only warm start before the mixed Stage~3 levels. Loss is displayed on a logarithmic axis.}
\label{fig:supp_teacher_curves}
\end{figure}

\paragraph{Metrics and plotting conventions.}
Figure~\ref{fig:supp_teacher_curves} shows success, mean episode return, and
critic fitting loss for the three teacher stages. Stage~1 success is measured
over completed episodes in each collection round. Stage~2 curves use
reset-balanced recent success within each active group; Stage~3 curves use
object-balanced recent success within each active group. Inactive groups
are not extended as if they were still being measured. Mean episode return
is the sum of recorded rewards over a completed episode, including the
absorbing-success tail. It is not an estimate of success probability.

The horizontal axes count PPO updates within each stage. Stage~1 has 194
updates, Stage~2 has 881, and Stage~3 has 136. Dashed lines mark curriculum
changes within a stage. Thin traces show raw recorded values; where a thicker
trace is present, it is a trailing seven-update mean, not an interpolation
between missing measurements. These are the actual training runs used for
the final checkpoint lineage, not averages over independent training seeds.
The Stage~2 run includes the continuation from one to four GPUs.

\begin{figure}[tbp]
\centering
\includegraphics[width=\textwidth,height=.72\textheight,keepaspectratio]{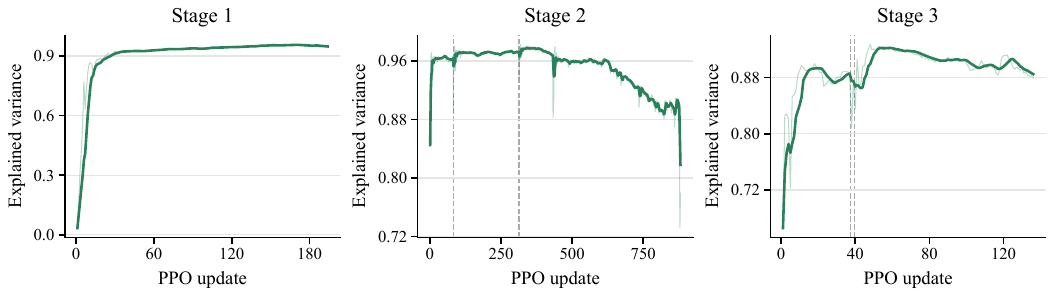}
\caption{Explained variance of the teacher critic during each stage. The target distribution changes as resets become more difficult. These curves complement value loss rather than implying a globally accurate value surface.}
\label{fig:supp_explained_variance}
\end{figure}

\paragraph{Why report more than reward and loss?}
A change in reset difficulty changes the state and return distributions.
A temporary decline in success at a curriculum transition therefore does
not by itself imply that the policy has lost a previously learned skill.
Group-wise success separates learning the new setting from retaining the
earlier one. PPO's policy-surrogate loss also changes with the freshly
collected advantages, so it is not a supervised prediction error that must
decrease monotonically. We prioritize the task success, return, and critic
loss curves, and add explained variance in
Figure~\ref{fig:supp_explained_variance}:
\begin{equation}
\mathrm{EV}=1-\frac{\operatorname{Var}(\hat G-V)}{\operatorname{Var}(\hat G)}.
\end{equation}
Higher EV indicates better variance prediction on the collected rollout;
it does not establish global accuracy of the critic outside its visited
state distribution. The KL target and hard limit are given in
Table~\ref{tab:supp_ppo} and provide additional optimizer diagnostics.

\begin{figure}[tbp]
\centering
\includegraphics[width=\textwidth,height=.72\textheight,keepaspectratio]{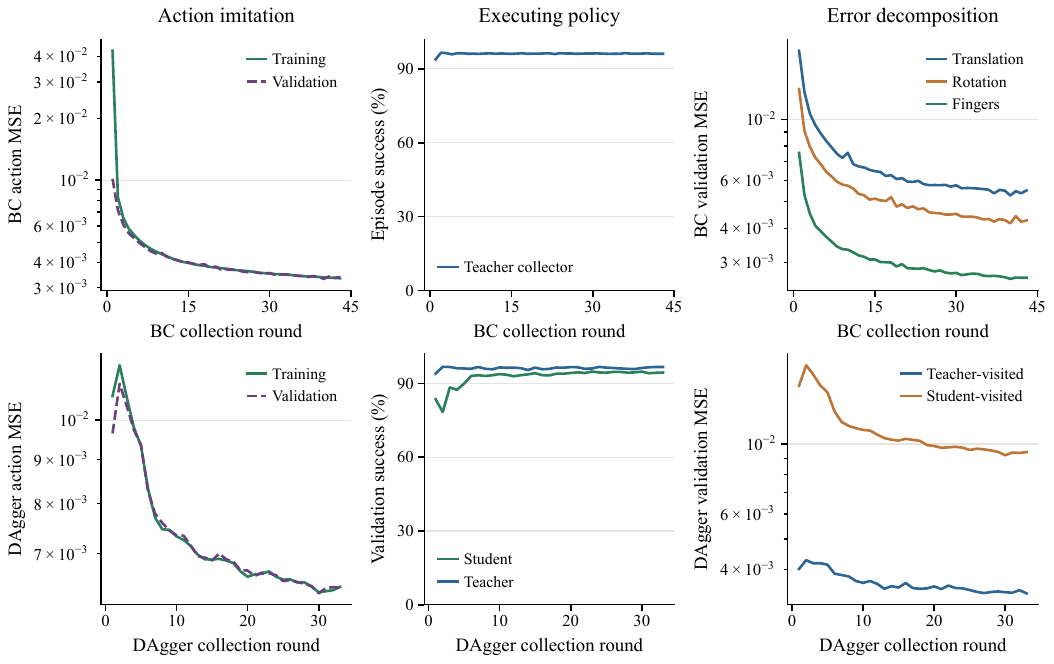}
\caption{Student distillation. Top: BC over 43 collection rounds. Bottom: DAgger over 33 rounds initialized from that BC checkpoint. Each collection is followed by 256 optimizer steps. Loss panels use logarithmic axes; all action errors are normalized-space MSE. The DAgger validation-success panel evaluates the policy actually executing each set of worlds.}
\label{fig:supp_student_curves}
\end{figure}

\paragraph{BC and DAgger diagnostics.}
Figure~\ref{fig:supp_student_curves} plots training and validation MSE in
the same normalized 18-dimensional action space. Training loss averages
the optimizer steps within a round; validation loss is evaluated after the
updates. Translation, rotation,
and finger errors average over their respective 3, 3, and 12 components.
The middle BC panel is explicitly the frozen teacher's collection success;
autonomous student success was not measured by those teacher-only rollouts.
During DAgger, the student and teacher execute separate validation worlds,
so their observed success rates can be reported alongside fitting errors.
The source-specific errors distinguish teacher-visited from student-visited
states. Validation worlds use held-out trajectories of training objects,
not held-out object identities; the main-paper asset evaluations measure
object generalization separately.

\end{document}